# Micro-Facial Expression Recognition in Video Based on Optimal Convolutional Neural Network (MFEOCNN) Algorithm

**S. D. Lalitha**

Assistant Professor, R.M.K Engineering College, Chennai, India
E-mail: sdl.it@rmkec.ac.in

**K. K. Thyagharajan**

Professor & Dean (Academic), Department of ECE, R.M.D Engineering College, Chennai, India
E-mail: kkthyagharajan@yahoo.com

Abstract: Facial expression is a standout amongst the most imperative features of human emotion recognition. For demonstrating the emotional states facial expressions are utilized by the people. In any case, recognition of facial expressions has persisted a testing and intriguing issue with regards to PC vision. Recognizing the Micro-Facial expression in video sequence is the main objective of the proposed approach. For efficient recognition, the proposed method utilizes the optimal convolution neural network. Here the proposed method considering the input dataset is the CK+ dataset. At first, by means of Adaptive median filtering preprocessing is performed in the input image. From the preprocessed output, the extracted features are Geometric features, Histogram of Oriented Gradients features and Local binary pattern features. The novelty of the proposed method is, with the help of Modified Lion Optimization (MLO) algorithm, the optimal features are selected from the extracted features. In a shorter computational time, it has the benefits of rapidly focalizing and effectively acknowledging with the aim of getting an overall arrangement or idea. Finally, the recognition is done by Convolution Neural network (CNN). Then the performance of the proposed MFEOCNN method is analysed in terms of false measures and recognition accuracy. This kind of emotion recognition is mainly used in medicine, marketing, E-learning, entertainment, law and monitoring. From the simulation, we know that the proposed approach achieves maximum recognition accuracy of 99.2% with minimum Mean Absolute Error (MAE) value. These results are compared with the existing for MicroFacial Expression Based Deep-Rooted Learning (MFEDRL), Convolutional Neural Network with Lion Optimization (CNN+LO) and Convolutional Neural Network (CNN) without optimization. The simulation of the proposed method is done in the working platform of MATLAB.

*Keywords: Adaptive median Filter, Micro-Facial Expression, Histogram of Oriented Gradients, Local binary pattern, Geometric features, Lion optimization, Convolution neural network.*

## 1. INTRODUCTION

Recognizing facial emotions using machine learning is considered as one of the challenging tasks in the recent years. For conveying the emotions and intentions some nonverbal methods like facial expressions, gestures etc., are used by human beings. Still it is a challenging task to identify the facial expressions by machine whereas, the expressions are quickly recognized by every individual without any delay or effort. For the real time applications, recognition of human emotions is encouraging in computer vision and image processing [1]. It is challenging but essential in the human-computer environment for the development of automatic recognition of facial emotions.

From the recent researches, it is clear that more maturely the recognition of facial expressions has grown. The recent research techniques on facial expression recognition are classified on the basis of data sources or the targets to be recognized. From the recognition target's point of view the facial expressions like neutral, disgust, fear, sad, anger, surprise and joy are recognized by many facial expression recognition methods [2]. Alternatively, to describe the facial expressions by AUs (action units), a system FACS (facial action coding) has been introduced. There are 46 AUs in the FACS system which is used for describing the fundamental facial expressions on the basis of activities of muscles [3].

Generally, from the data sources' point of view the methods of recognition of facial emotions are classified into image sequence (video) based techniques and static image based techniques. For extracting the facial features, a typical method is estimating the optical flow in image sequence based methods [4]. Conversely, it is more difficult to recognize the facial emotions from static images because there is no availability of temporal information in the static images. Initially, a neutral face is considered as reference face and it should be recognized in the image sequence based techniques [5]. Then depending on the difference between the input and the reference image the recognition is performed. However, it is challenging for recognizing the facial emotions of the input images if the reference image is not recognized correctly. Hence, for the emotion recognition static images are often used by psychologists [6].

Among the facial expression recognition techniques, appearance based methods are the most popular. For avoiding the problems in landmark or feature point detection, the appearance images are processed and the features are extracted as two-dimensional holistic patterns [7]. In order to classify the holistic patterns, these methods are generally based on LDA (Linear discriminant analysis). Anyhow, in high dimensional data the methods based on LDA are suffered from small-sample size problem [8].

Due to the variation in the faces of one person to another automatic recognition of facial expressions is difficult [9]. Though, the recognition is performed with some constraints particular to culture, various issues such as presence of glasses, facial hair make the recognition task more complex. Orientation of face and variation in the input image space are the other challenges included in the recognition process. Because of this, in the images a search for a fixed pattern is disabled. Owing to the camera angle, there may be a difference in the poses of faces. The faces may be towards frontal or non-frontal. Some facial features are obscured because the faces are at different angles [10].

To apply on the input images, various preprocessing methods are needed which have good insensitivity for rotation, scaling and translation of the head. Recently, geometrical information or local spatial analysis are used as facial features by feature based recognition methods. For the robust classification of facial expressions, an important step discriminant analysis) and LDA (linear discriminant analysis) were combined in the RDA. Ill-posed and small sample size issues suffered from is automatic localization of facial points [11].

LDA and QDA were solved by RDA by using various regularization techniques. Also, in the RDA, for the estimation of optimal parameters PSO (particle swarm optimization) method was used. The results from this approach have demonstrated that the

Anyhow, the environmental conditions like lightning to be depended by the performance of extraction of facial point in the real time applications like robotics. Therefore, the facial point can be identified inaccurately if there is nonuniform illumination. Hence, the facial expression's recognition rate cannot be expected highly. This issue makes the feature extraction process more challenging [12]. Before the process of feature extraction, image processing techniques like Rank Normalization, Histogram Equalization and DCT normalization are applied for the compensation of illumination variation in the input image. Considering these issues, a new method has been proposed so as to tackle these issues.

The outline of the paper is given as follows, a brief review of researches in facial expression recognition is provided in section 2. In section 3, the contribution of the proposed method is presented. The proposed

Micro-Facial expression recognition is presented in section 4. In section 5, the proposed performance evaluation and comparative analysis is presented. Finally the conclusions are summed up.

## 2. RELATED WORK

As facial emotion recognition (FER) is one of a challenging tasks, to cope with the problems in recognizing the facial emotions (Arora et.al, 2018) have presented an effective framework. For encoding the facial components into features the gradients have been applied. For the facial emotion recognition, by using the random forest classifier the presented model was trained. Further, by the testing process it has been combined to classify the emotions. The presented method has robust feature extraction techniques. Results have exposed that the presented approach has outperformed the existing approaches in terms of recognition accuracy, false rejection rate and false acceptance rate.

In autism spectrum disorder (ASD) a key deficit field is recognition of facial emotions. In youth with ASD, for examining the trajectories of facial emotion recognition (Rosen et.al, 2016) have presented a study in which the repeated measurements over 18 weeks are recorded and the effects of externalizing and internalizing symptoms are evaluated. Even for the difficult stimuli, the errors in the facial emotion recognition are reduced over time which was revealed by the analysis of Hierarchical Linear Modeling. The externalizing symptoms have attenuated the improvement of FER whereas, the internalizing symptoms have enhanced the improvement of FER.

A boosting approach based on RDA (regularized discriminant analysis) was developed and its application in FER was discussed by (Lee et.al, 2012). QDA (quadratic discriminant analysis) and LDA (linear discriminant analysis) were combined in the RDA. Ill-posed and small sample size issues suffered from LDA and QDA were solved by RDA by using various regularization techniques. Also, in the RDA, for the estimation of optimal parameters PSO (particle swarm optimization) method was used. The results from this approach have demonstrated that the facial emotions were recognized robustly and accurately. When compared with the existing techniques the proposed approach outperformed them.

In real-world natural situations, for robustly recognizing the facial emotions (Shojaeilangari et.al, 2015) have proposed a system with ESL (Extreme Sparse Learning). The proposed method has the capability for learning both the non-linear classification and dictionary model. To provide accurate classification by the proposed method, the reconstruction property of the sparse representation has been combined together with the ELM's (extreme learning machine) discriminative power. On both the spontaneous facial emotion and acted databases, the recognition accuracy of state-of-the-art methods has been achieved by the proposed method.

In the recognition of facial emotions, still, there are some open challenging issues for simple and efficient feature selection and classification. During the classification process, for reducing the interclass pixel mismatch problem, a simple and effective FER approach has been developed by (Krestinskaya et.al, 2017). For removing the intensity offsets with Min-Max metric which is able to suppress the feature outliers, pixel normalization application has been included in the proposed approach. Though this method has been able to overcome many of the existing methods, the memory requirement for implementing the proposed method is very large.

(Sebe et.al, 2002) have proposed a method for recognizing the emotions from the facial expressions in the video sequences. They have used a classifier called Cauchy Naive Bayes in which the model distribution is Cauchy distribution. To select the best assumption of model distribution, a framework has been provided. Results have shown that the Cauchy distribution assumption has provided a much better outcome than Gaussian distribution assumption.

For the automatic recognition of emotions, an approach has been proposed by (Dhall et.al, 2011). To encode the appearance and shape information, the LPQ (local phase quantization) and PHOG (pyramid of histogram of gradients) have been extracted in this approach. From the face tracking based on CLM (constraint local model), the normalized shape vectors have been derived. Then, k-means clustering has been applied to select the key frames. The proposed method has performed better than the baseline result.

## 3. CONTRIBUTION

Our contribution in this paper is, we proposed an effective Micro-Facial expression recognition in video on the basis of optimal convolution neural network.

The proposed system performs the following procedure,

*a)* For Micro-Facial expression recognition, a novel algorithm called MFEOCNN is proposed. To identify the facial expression, efficient features are selected optimally by means of Modified lion optimization (MLO) method.
*b)* The proposed Micro-Facial expression recognition is done by convolutional neural network, which is recognized the facial expression in to relevant expression and irrelevant expression.

## 4. PROPOSED METHOD

In this paper, human facial expression recognition process is developed using three steps such as, i) Preprocessing ii) Extraction of features iii) Optimal feature selection and Micro-Facial expression recognition (MFEOCNN). Here the proposed method considering the input dataset is CohnKanada (CK+) dataset. Initially the input video is converted into number of frames, and then each frame is given to the input for preprocessing stage. The main intension of preprocessing stage is getting image as clear with normalized intensity, uniform size and shape. For preprocessing, an adaptive median filter is considered by the proposed system. Then the features are extracted from the resultant output. To extract the appearance variations of a testing image-spatial analysis micro facial features are utilized. In a sequence of facial images, to recognize the expressions motion information features are utilized. The proposed method considers the Histogram of Oriented Gradients (HOG) features, the local binary pattern (LBP) features and the Geometric features. Once the features are extracted from the input image they are given to further process. Optimal feature selection and Micro-Facial expression recognition is the final step of the developed method. Here the extracted features are selected optimally by means of MLO.

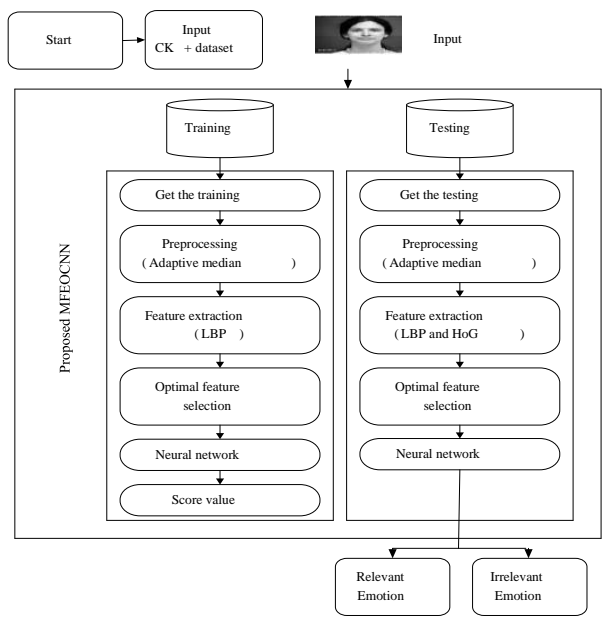

**Fig.1. Flow diagram of proposed Micro-Facial expression recognition in video**

Next the optimal features are utilized to recognize the Micro-Facial expressions. For recognizing the facial expression the suggested method utilizes CNN (convolutional neural network). The facial expression is effectively recognized by the proposed method based on CNN [23]. The overall flow diagram of the proposed recognition method is shown in fig.1.

Three Important steps in the proposed technique are,
- Preprocessing
- Feature extraction
- Optimal feature selection and Micro-Facial expression recognition

## 4.1 Preprocessing

Initial stage of proposed Micro-Facial expression recognition is preprocessing. Preprocessing strategies predominantly went for the improvement of the face image without adjusting the data contained in an image. For the preprocessing, an adaptive median filter is utilized. This compares each pixel with its neighbor pixels to classify the noise pixels. The pixels which are not aligned structurally and are different from their neighbor pixels are denoted as impulse noise. Then the median pixel value is assigned to these noise pixels.

Let $X_{ij}$ be the pixel value of input image, $X_{min}$ and $X_{max}$ indicate the minimum and maximum pixel value of the image. $X_{med}$ represents the median value. $W_{crnt}$ be the current window size, $W_{mnin}$ and $W_{max}$ represents the minimum and maximum widow size. $W_{crnt}$ starts with the $W_{min}$. The procedure of adaptive median filter is given beneath.

---

***Procedure of adaptive median filter:***

Input: Input image.
Output: Preprocessed image. Start
Capture Image.
Preprocess the image.
Identify minimum, maximum and median value of the image.
Initialize the current window size, minimum and maximum window size.

Level A:

If $X_{min} < X_{med} < X_{max}$ then $med \square \text{Im} \, pulsenoise$, so the algorithm pass to level B.
Else, the window size is increased, in which replace the pixel value with median value and repeat level A.

Level B:

If $X_{min} < X_{ij} < X_{max}$ then $med \square \text{Im} \, pulsenoise$, so the pixel value is unchanged.
Else pixel is either equal to $X_{max}$ or $X_{min}$ then replace the pixel value with median value from level A.
Restore the image pixel.
End
Stop

---

Based on that, the noises from the input image or frames are removed and it will prepare the input frame for the next process.

## 4.2 Feature extraction

Recognizing the facial expressions is a very challenging. In order to overcome this difficulty, the implemented method has to identify the uniqueness of each image under various expressions. In the proposed system, the feature values are calculated from the Neutral image as well as from the expression image. In our examination, the feature extracted from the preprocessed output is,

- Geometric Features
- HOG features
- LBP features

### 4.2.1 Geometric Features

In geometric based feature extraction [27] strategy, points which portray the geometric data related with facial features for example, eye, eyebrow, and mouth are set apart on the face. The proposed strategy needs to label each image with a lot of landmark points to discover the varieties in the face by ascertaining the deviation esteems. To discover the landmark points, pixels between the neutral image and the distinctive face responses of a specific individual is looked at. By utilizing the Euclidean distance, the contrast between the pixels is determined. The estimation of Euclidean distance is as per the following equation

$$d(x,y) = d(y,x) = \sqrt{(y_1 - x_1)^2 + (y_2 - x_2)^2 + \ldots + (y_m - x_m)^2} \quad \ldots\ldots (1)$$

After calculating the Euclidean distance, we need to apply the landmark points. The landmark points are connected as pursues: in the event that the Euclidean distance for a specific pixel is zero, at that point we need to prohibit that pixel. On the off chance that the Euclidean distance for a specific pixel gets any esteem, at that point we need to stamp that pixel as landmark point. From the input image the feature values are extracted based on the above process, then the extracted features are fed to the input for the next process.

### 4.2.2 Local Binary Pattern

The LBP [25] feature extraction strategy is a basic and non-parametric technique that depicts spatial data of the pixels in regards to their neighbor pixels. This is finished by appointing a label to every pixel utilizing the below equation,

$$LBP = \sum_{i=0}^{m-1} S(G_n - G_c) 2^i \quad \ldots\ldots (2)$$

Where, H G and V G represents the horizontal and vertical gradients.

### 4.2.3 Histogram of Oriented Gradients

The HOG [26] feature extraction strategy utilizes local gradients to depict the shape of an object. To this end, the horizontal and vertical gradients of a given picture are determined first. At that point, the magnitude (M) and the direction (D) of every pixel's gradient are shown in below equation.

$$M = \sqrt{G_H^2 + G_V^2} \quad (3)$$

$$D = \arctan \frac{G_V}{G_H} \quad (4)$$

the acknowledgment issue. Here the traditional lion algorithm [22] is modified by the updation function of PSO [24] algorithm. The proposed modified lion optimization algorithm based feature selection is clarified.

## 4.3 Optimal Feature Selection and Micro-Facial Expression Recognition

Feature selection in itself is one of the vital research territories in the space of machine learning. The principle of selecting features is getting the most optimal features which provide assistance in enhancing the recognition accuracy and lessening computational overhead, resource request and storage space necessity. In the process the most persuasive features get chose so that the user can be able to interpret the connection between the features and classes. In this paper, MLO method is proposed to choose the optimal features for the acknowledgment issue. Here the traditional lion algorithm [22] is modified by the updation function of PSO [24] algorithm. The proposed modified lion optimization algorithm based feature selection is clarified.

### 4.3.1 Modified Lion Optimization (MLO) Algorithm

MLO is a Meta heuristic algorithm which is based on the population. Meta heuristic algorithms can produce distinctive answers for the issue in each run. Lion has one of a kind social conduct so it is the most grounded warm blooded animal on the planet. Lions have two sorts of social conduct: residents and nomads. Lion can switch the kind of social association, residents may progress toward becoming nomads and the other way around. Residents live in groups called as pride, in which the resident males and females are taken care of conceive an offspring. The second authoritative conduct is nomads, who move periodically either in sets or independently. Sets are more observed among the related males who have been avoided from the maternal pride. The MLO comprises of following steps,

- Evaluate fitness value for each lion
- Perform hunting operation and moving towards safe place for each lion
- Perform roaming operation for each lion
- Perform mating operation for each lion
- Perform updation operation using velocity updation

These steps are repeated until some stopping condition is met.

*STEP 1: Fitness evaluation:* By calculating the objective function, the fitness value of each lion L is calculated. The fitness function is shown in below,

Fitness = Max Accuracy   …. (6)

*STEP 2: Hunting operation:* The hunters are arbitrarily separated into three sub groups.
The group with most elevated fitness is considered as the middle and the other groups. are considered as right and left wings. All through chasing, if a seeker enhances its own fitness, prey (Pr) will escape from the seeker and then the new position of the prey is,

$$P_r^{'} = P_r + rand(0,1) \times \%I \times (P_r - h) \quad \ldots (7)$$

Where, $P_r$ Indicates the current position $h$ Indicates the hunter, $\%I$ indicates the improvement percentage of fitness.

The new places of the hunters which are having a place both left and right wings,

$$h^{'} = \begin{cases} rand(h, P_r), & h < P_r \\ rand(P_r, h), & h > P_r \end{cases} \quad \ldots.. (8)$$

*STEP 3: Moving towards safe place:*
The female lion's new position is,

$$P'_F = P_F + 2d \times rand(0,1)\{SP_1\} + U(-1,1) \times \tan(\theta) \times D \times \{SP_2\} \quad \ldots (9)$$

$P_F$ Indicates the female line position

The distance between the chosen point by tournament selection and the female lion is indicated as D,

$\{SP_1\}$ Indicates a vector for which, the starting point is the female lion's previous location and its moving direction is towards the selected point,

$\{SP_2\}$ And $\{SP_1\}$ are perpendicular

A low number of achievement demonstrates that the lions are swinging around the optimum point without critical enhancement. Additionally, a high number of achievements shows that they have met at a point which is far away from the optimum point. With the goal that the measure of the competition is assessed utilizing the achievement values.

$$T_{S_j}(SU) = \sum_{i=1}^{N} SU(i,t,G) \quad j = 1,2,\ldots N \quad (10)$$

$$SU(i,t,G) = \begin{cases} 1 & Bst^t_{i,G} < Bst^{t-1}_{i,G} \\ 0 & Bst^t_{i,G} = Bst^{t-1}_{i,G} \end{cases} \quad (11)$$

$Bst^t_{i,G}$ Indicates the best position until iteration t, $T_{S_j}(SU)$ Indicates the number of lions in the pride j which has an enhancement in the fitness for the last iteration,

For each iteration, the tournament size is unstable, it implies that when SU esteem diminishes, $T_{S_j}(SU)$ is expanded and it prompts assorted variety. With the goal that the competition estimate is determined utilizing beneath equation,

$$T_{S_j} = \max\left(2, ceil\left(\frac{T_{S_j}(SU)}{2}\right)\right) \quad j = 1,2,\ldots G \quad (12)$$

*STEP 4: Roaming operation:* Roaming is a local search which helps the lion optimization algorithm (LOA) to look for a solution to enhance it. By n units, the lion moves towards the chosen region, where n is a random number with uniform distribution.

$$n \sim U_d(0, 2 \times D) \quad (13)$$

The distance between the selected area of territory and the position of male lion is indicated as D.

Also, in the search space the nomad lions are moving chaotically. The new positions of nomad lions are given as pursues,

$$L_{ij} = \begin{cases} L_{ij} & \text{if } rand_j > prob_i \\ RAND_j & \text{otherwise} \end{cases} \quad (14)$$

The probability value is denoted as $prob_i$, the random vector is indicated as $RAND_j$, the uniform random number between 0 and 1 is denoted as $rand_j$ and the current position of nomad is represented as $L_{ij}$.

For every nomad lion, the probability value is independently calculated by using the below equation,

$$prob_i = 0.1 + \min\left(0.5, \frac{(N_{O_i} - Bst_{N_O})}{Bst_{N_O}}\right) \quad i = 1,2,\ldots M \quad (15)$$

Where,

$Bst_{No}$ - represents the optimum delay of nomad lion

$N_{Oi}$ - represents the delay of i<sup>th</sup> lion
The number of nomad lions is represented as M.

*STEP 5: Mating operation:* Mating is the vital one that guarantees the lions survival. A direct combination of parents is the mating operator to produce two new offspring. After choosing the male lions and female lion for mating, the new cubs are generated.

$$Offspring_j 1 = \chi \times F_j + \sum \frac{1-\chi}{\sum_{i=1}^{N_R} SU_i} \times M_j \times SU_i \quad (16)$$

$$Offspring_j 2 = (1-\chi) \times F_j + \sum \frac{\chi}{\sum_{i=1}^{N_R} SU_i} \times M_j \times SU_i \quad (17)$$

Here, j is the dimension. If the male lion *i* is selected for mating, then $SU_i = 1$, otherwise $SU_i = 0$.

A random number with uniform distribution with standard deviation 0.1 and mean 0.5 is denoted as ☐ and the resident male lions count in a pride is denoted as $N_R$.

In this lion algorithm, the information between the genders are shared by mating, while the new cubs get character from the two genders.

*STEP 6: Velocity Updation:* The position of lion is updated in the search space using the below equation,

$$Lp_i^{New} = Lp_i + V_i^{New} \quad (18)$$

$$V_i^{new} = V_i + \varphi_1.r_1.(pbest_i - Lp_i) + \varphi_2.r_2.(gbest_i - Lp_i) \quad (19)$$

$\Phi_1, \Phi_2$ - Learning rate $r_1, r_2$ - Random numbers in the range of [0, 1].

$V_i$ - indicates the current velocity.

$Lp_i$ - indicates the current lion position
The overall flow of the proposed MLO algorithm is given in fig.2.

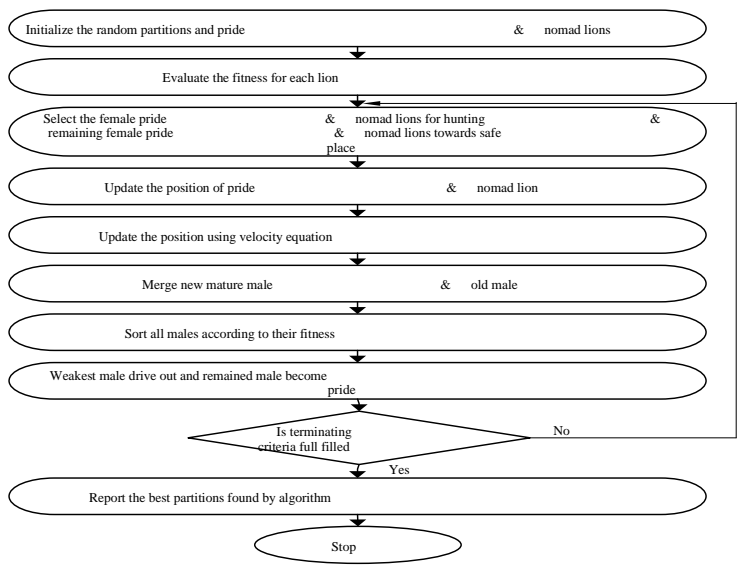

**Fig.2. flow chart of the modified lion optimization algorithm**

From the above process the optimal features are selected for the proposed MFEOCNN. By utilizing this two updation process, it will conquer the weaknesses of the individual execution. Also, in a shorter computational time it has the benefits of rapidly focalizing and effectively acknowledging with the aim of getting an overall arrangement or idea. Then the output of this process is given to the recognition process. The clear explanation of the proposed recognition approach is described in further section.

*Recognizing Micro-Facial Expression using Convolution Neural Network:* The proposed framework is developed by using a CNN with three layers which can distinguish and recognize the images into predefined classes. Here the strategy is thinking about the contribution as the optimal features. In CNN, the optimal features are directly given to the input feature map. There is no need to find the feature map in CNN. So, when compared to the traditional CNN the developed approach reaches the minimum processing time. CNN is typically made by a set out of layers that can be assembled by their functionalities. It is variations of MultiLayer Perceptron (MLPs) which are enlivened from biology. Convolution neural network can be connected to a wide assortment of computational tasks. The network comprises of three sorts of layers specifically convolution layer, output layer and sub sampling layer. The proposed CNN is appeared in fig.3.

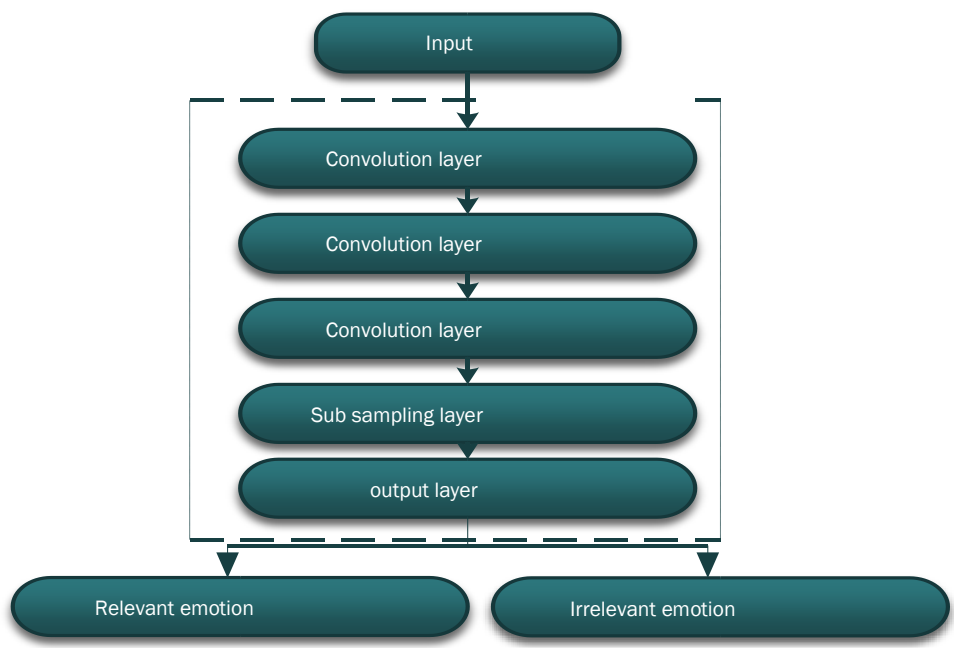

Fig.3. proposed convolution neural network

*STEP 1: Convolution layer:* In the CNN network, the convolution layer is the primary layer. It comprises of a convolution mask, bias terms and a function expression. Together, these create yield of the layer. The structure of convolution layer is shown in fig.4. Fig.4 demonstrates a 5x5 convolution mask that perform convolution over a 32x32 input feature map. The resultant yield is a 28x28 matrix. At that point bias is included and sigmoid function is connected on the matrix.

*STEP 2: Sub sampling layer:* This layer comes after the convolution layer. As in the convolutional layer, it has the same number of planes. The motivation behind this layer is to decrease the measure of the feature map. It isolates the input blocks of 2x2 and performs averaging. Sub sampling layer protects the relative data among features and not the definite connection. The structure of sub examining layer is appeared underneath,

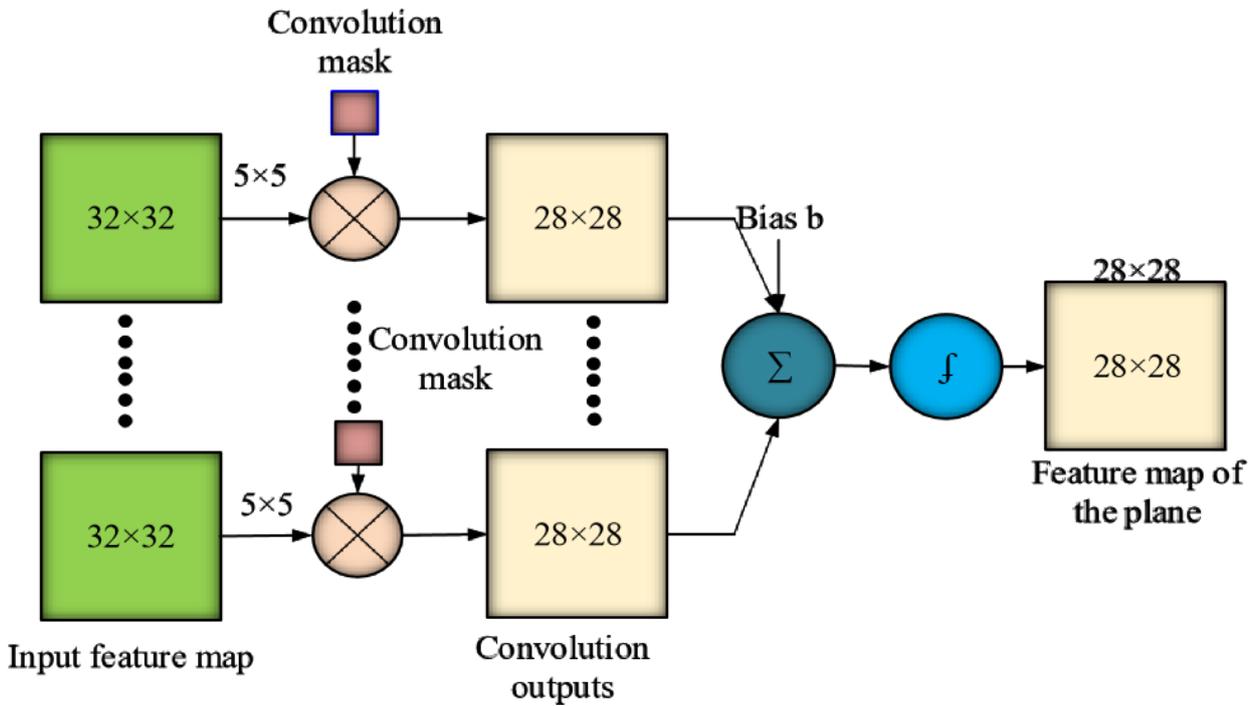

Fig.4. Convolution layer structure

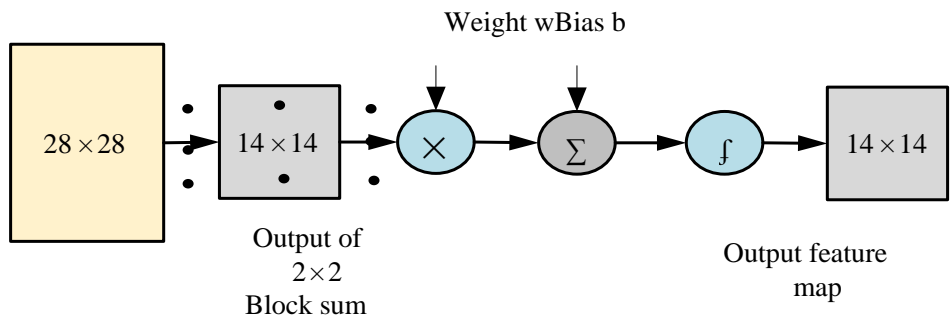

Fig.5. Sub sampling layer structure *Output layer*

This layer plays out the recognition of the facial expression dependent on the features separated by the past convolutional layers. In completely connected layer, each neuron is associated with each neuron of past layer. A soft max function is utilized to change over the yield of neural network into likelihood for each class. In view of the above procedure the facial expression is recognized as relevant emotion or irrelevant emotion.

V. RESULT AND DISCUSSION

The proposed Micro-Facial expression recognition in video based on optimal CNN is done on the working platform of MATLAB (version 2017a). The proposed system is implemented in a windows machine with the configurations Intel (R) Core i5 processor, 3.20 GHz, 4 GB RAM, and the platform of operating system is Microsoft Window7 Professional. In the proposed technique, adaptive median filter is utilized for preprocessing. From the preprocessed image the LBP, HOG and geometric features are extracted. Then the optimal feature selection and MicroFacial expression recognition is done by Modified Lion Optimization (MLO) algorithm and Convolution neural network (CNN). Here the proposed method considering the input dataset is CK+ dataset [28]. The experimental results are shown in the below sections. The sample images of the input dataset is shown in fig.6.

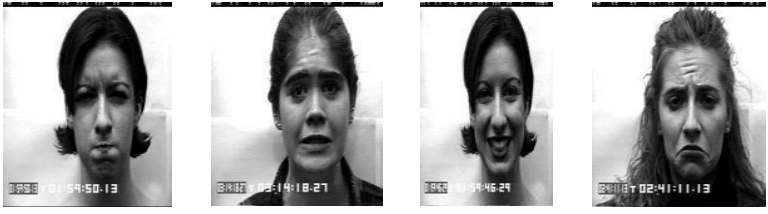

Anger    Fear    Happy    Sadness

**Fig.6. the sample input image of frame from CK+ dataset**

The details of simulation parameters utilized for the proposed method are given in Table 1.

**Table 1: Details of evaluation parameters**

| Parameters | Value |
|---|---|
| Data Set | Cohn-Kanada (CK+) dataset |
| No of Persons | 123 persons |
| No of video sequence | 593 video sequence |
| Image Resolution | 640×480 and 640×490 |
| Tool Used | Matlab 2017 a |

## 5.1 Performance Metrics

For the analysis of proposed recognition method, various performance metrics are evaluated. They are calculated by using TP (true positive), FP (false positive), TN (true negative) and FN (false negative) values with the option of pixel differences. The proposed method's performance is estimated by means of Specificity, Sensitivity, Accuracy, F-measure, Recall, Precision and Mean Absolute Error (MAE) [29].

*Mean Absolute Error*

MAE is the difference between two samples. More formally, MAE of a given data set is evaluated as follows:

$$MAE = \frac{1}{N}\sum_{i=1}^{N}(b_i - a_i) \qquad (20)$$

$a_i$ represents the regular of apparent expression annotations $b_i$ represents the predicted value $N$ represents the number of images.

*Precision*

The precision determines the number of pixels in the image is recognized as positive which are actually positive,

$$precision = \frac{TP}{TP + FP} \qquad (21)$$

*Recall*

It is the percentage of positives recognized correctly. It is also equal to Sensitivity and also it is called as Detection Rate.

$$recall = \frac{TP}{TP+FN} \quad (22)$$

*F-Measure*

F-measure is a weighted harmonic mean of precision and recall.

$$FMeasure = 2\frac{precision \times recall}{precision + recall} \quad (23)$$

*Sensitivity*

Sensitivity is the measure of how exactly recognition is done for positive results.

$$Sensitivity = \frac{TP}{FN+TP} \quad (24)$$

*Specificity*

Specificity is the measure of how exactly recognition is done for negative results.

$$Specificity = \frac{TN}{TN+FP} \times 100 \quad (25)$$

*Accuracy*

It is defined as the percentage of input images which are detected accurately.

$$Accuracy = \frac{TP+TN}{TP+FP+TN+FN} \times 100 \quad (26)$$

Facial expression recognition process is tested with different datasets of input face and feature points, and the future result of the proposed System has been shown below,

## 5.2 Analysis of proposed performance

The analysis of proposed performance is evaluated by accuracy, specificity, sensitivity, F-measure, recall and precision. The experimental outcomes of the proposed approach help to analyze the efficiency of the proposed Micro-Facial expression recognition. The subsequent table shows the analysis of the proposed performance.

Table.2 Performance analysis of proposed MicroFacial expression recognition

| Methods | Accuracy | Recall | F-Measure | Precision | Sensitivity | Specificity |
|---|---|---|---|---|---|---|
| MFEOCNN (CNN+MLO) | 0.992 | 0.972 | 0.972 | 0.972 | 0.972 | 0.995 |

When analyzing table.2, the proposed Micro-Facial expression recognition achieves the recognition accuracy value as 0.992. The proposed specificity value of MicroFacial expression recognition is 0.995. Sensitivity, FMeasure, recall and precision of the proposed method achieve the same value as 0.972. The efficiency of the proposed approach is analyzed and the results are compared with the existing algorithm in the following section.

## 5.3 Effectiveness of the proposed Micro-Facial expression recognition

In this section, the effectiveness of the proposed method is compared with the existing methods. Here we are considering the existing algorithms CNN and traditional lion optimization (CNN+LO), Micro facial expression based on Deep Routed learning (MFEDRL) [29] and CNN without optimization. The results are provided in table.3,

**Table.3 Effectiveness of proposed MFEOCNN**

| MEASURES (%) | Proposed MFEOCNN (CNN +MLO) | Existing MFEDRL | Existing CNN +LO | Existing CNN |
|---|---|---|---|---|
| Accuracy | 0.992 | 0.98 | 0.986 | 0.984 |
| F-Measure | 0.972 | 0.93 | 0.952 | 0.944 |
| Recall | 0.972 | 0.93 | 0.952 | 0.944 |
| Precision | 0.972 | 0.93 | 0.952 | 0.944 |
| Sensitivity | 0.972 | 0.93 | 0.952 | 0.944 |
| Specificity | 0.995 | 0.99 | 0.992 | 0.944 |

As shown in table.3, the proposed MFEOCNN approach attains 99.2% of accuracy value but the existing methods attain 98% for MFEDRL, 98.6% for CNN+LO and 98.4% for CNN, which are minimum when compared with the developed approach. Because of the optimal feature selection from the input image, the implemented MFEOCNN achieves maximum accuracy value. Here same values are attained for recall, F-measure, precision and sensitivity of the developed MFEOCNN approach. The proposed method's specificity value is 99.5% which is maximum value when compared with the other methods. The graphical depiction of the comparison of sensitivity, specificity and accuracy by using various classifiers is shown in fig.7. Likewise, the graphical representation of the comparison of F-measure, recall and precision by using various classifiers is shown in fig.8\

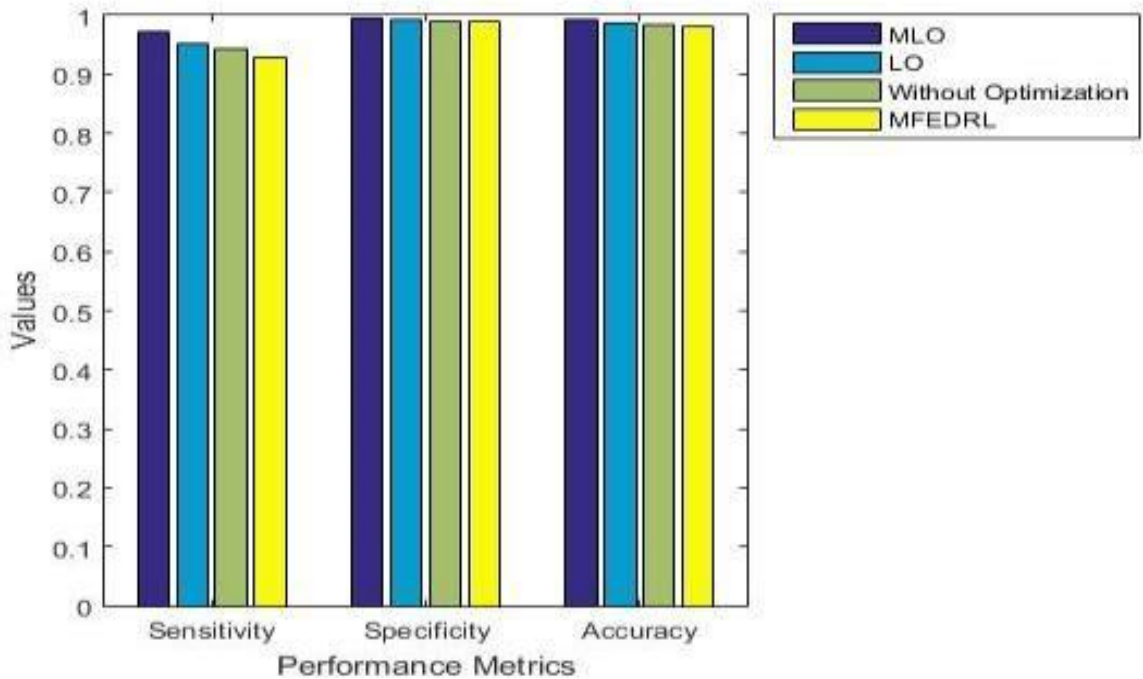

**Fig.7. the comparison of sensitivity, specificity and accuracy of different classifiers**

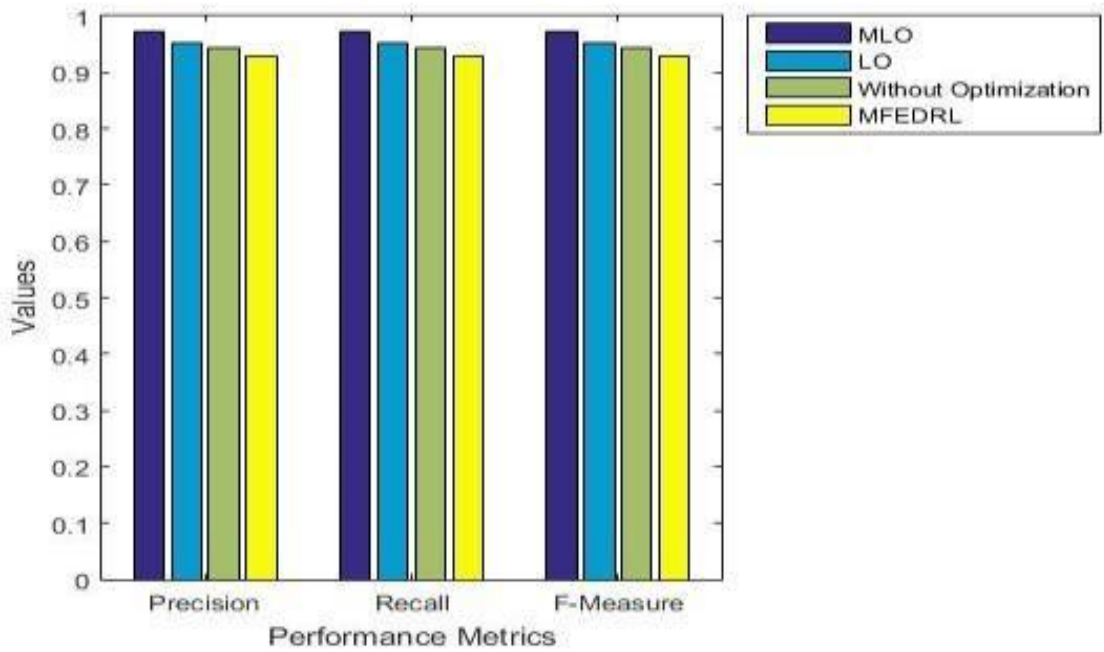
**Fig.8. the comparison of F-measure, recall and precision of different classifiers**

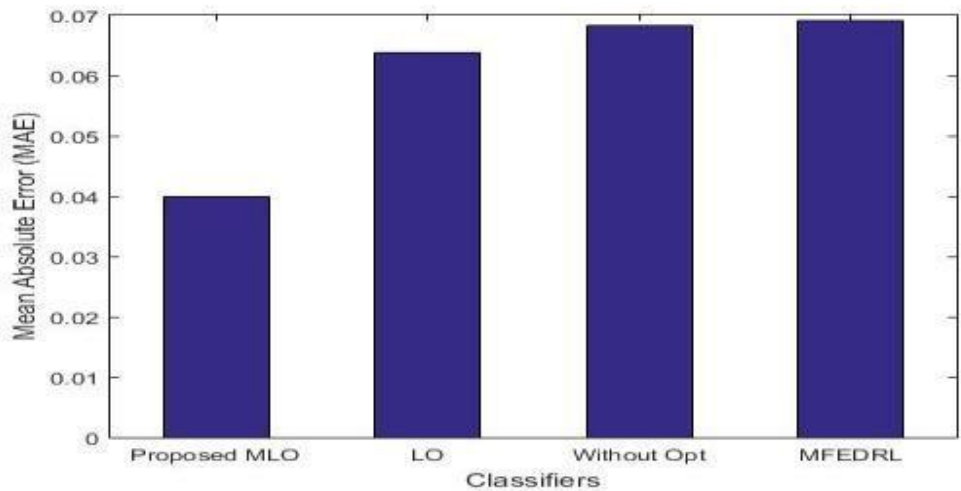
**Fig.9. Mean Absolute Error (MAE) for various Methods**

The comparison of Mean Absolute Error (MAE) of the various methods is shown in fig.9. The proposed microfacial Expression recognition MFEOCNN increases the percentage of Accuracy with reduced error level (MAE) when compared with existing methods. When analysing the above figure, MFEOCNN approach attains the minimum MAE value when compared with other methods. Here, MFEOCNN method gets the MAE value as 0.0398 but the existing method CNN+LO gets the MAE value as 0.063 whereas, CNN and MFEDRL get the MAE value as 0.068 and 0.069 respectively. From the graphical outcomes, it is clear that the developed MFEOCNN (CNN+MLO) computes the correct and incorrect estimation with the maximum accuracy and minimum MAE value. The comparison of overall processing time of the developed and existing methods is depicted in fig.10.

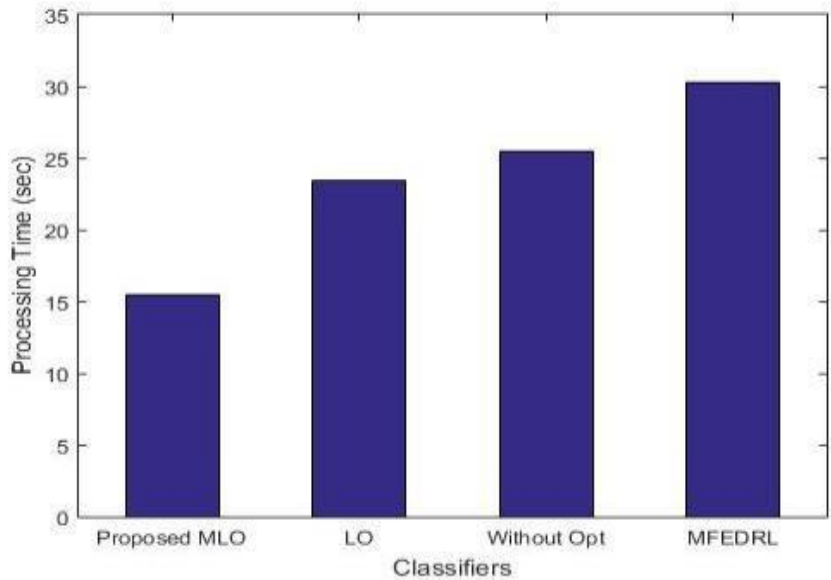

**Fig.10. The overall processing time for various Methods**

From fig.10, it is depicted that the proposed method reaches the minimum processing time when compared with the existing methods. Here MFEOCNN method reaches the processing time as 15.45 sec, which is minimum when compared with other methods. The recognition performance of the proposed MFEOCNN method is compared with various research papers and the results are provided in table 4.

**Table.4 Comparison of recognition performance of various papers**

| Methods | Recognition Accuracy |
|---|---|
| Proposed MFEOCNN | 99.2% |
| CDMML [20] | 96.6% |
| LEMHI-CNN [21] | 83.2% |

Here all the papers considered the input is CK+ dataset. For comparison, the proposed method considers the existing method is Collaborative Discriminative Multi-Metric Learning (CDMML) [20] and Local Enhanced Motion History Image and CNN [LEMHI-CNN]. When analysing the above table.4, MFEOCNN method attains the maximum value when compared with other methods. From the results, it is clearly depicted that the MFEOCNN achieves better recognition accuracy and minimum MAE and processing time on comparing with other approaches.

## 6. CONCLUSION

The optimal micro-facial expression recognition from the CK+ dataset is developed in this paper. The analysis of the performance of proposed method is done by the performance metrics accuracy, specificity, sensitivity, FMeasure, recall and precision. The experimental outcomes show the effectiveness of the Micro-Facial expression recognition by recognizing the emotion as correct and incorrect estimation with the minimum MAE value. Results predicted the level of error, and also it proved that the micro-facial Expression recognition accuracy is better when compared with other methods. Here, the developed MFEOCNN attains 99.2% recognition accuracy value. From the results, it is depicted that the MFEOCNN method recognizes better than the others. In future, to ensemble the need of different security models, facial expression framework can be focused, for example, legislative private security breaks, criminal location etc. Furthermore, depth information, optimal fusion of color and more profundity about the face classification issue can be examined.